\documentclass[10pt,twocolumn,letterpaper]{article}

\usepackage{cvpr}              

\usepackage[dvipsnames]{xcolor}

\usepackage{multirow}
\usepackage{xcolor}
\usepackage[normalem]{ulem}
\definecolor{cvprblue}{rgb}{0.21,0.49,0.74}
\usepackage[pagebackref,breaklinks,colorlinks,citecolor=cvprblue]{hyperref}
\usepackage{xcolor} 

\usepackage[capitalize]{cleveref}
\crefname{section}{Sec.}{Secs.}
\Crefname{section}{Section}{Sections}
\Crefname{table}{Table}{Tables}
\crefname{table}{Tab.}{Tabs.}

\newcommand{\Ours}{S'More }
\newcommand{\Ourstight}{S'More}

\def\paperID{338} 
\def\confName{CVPR}
\def\confYear{2024}

\title{Instantaneous Perception of Moving Objects in 3D}

\author{Di Liu$^1$ \quad Bingbing Zhuang$^3$ \quad Dimitris N. Metaxas$^1$ \quad Manmohan Chandraker$^{2,3}$  \vspace{+0.3em} \\
$^1$Rutgers University~~~$^2$University of California, San Diego~~~$^3$NEC Labs America\vspace{-0em} \\
}

\begin{document}
\twocolumn[{%
\renewcommand\twocolumn[1][]{#1}%
\maketitle
\vspace{-28pt}
\begin{center}
    \centering
\includegraphics[width=0.91\linewidth]{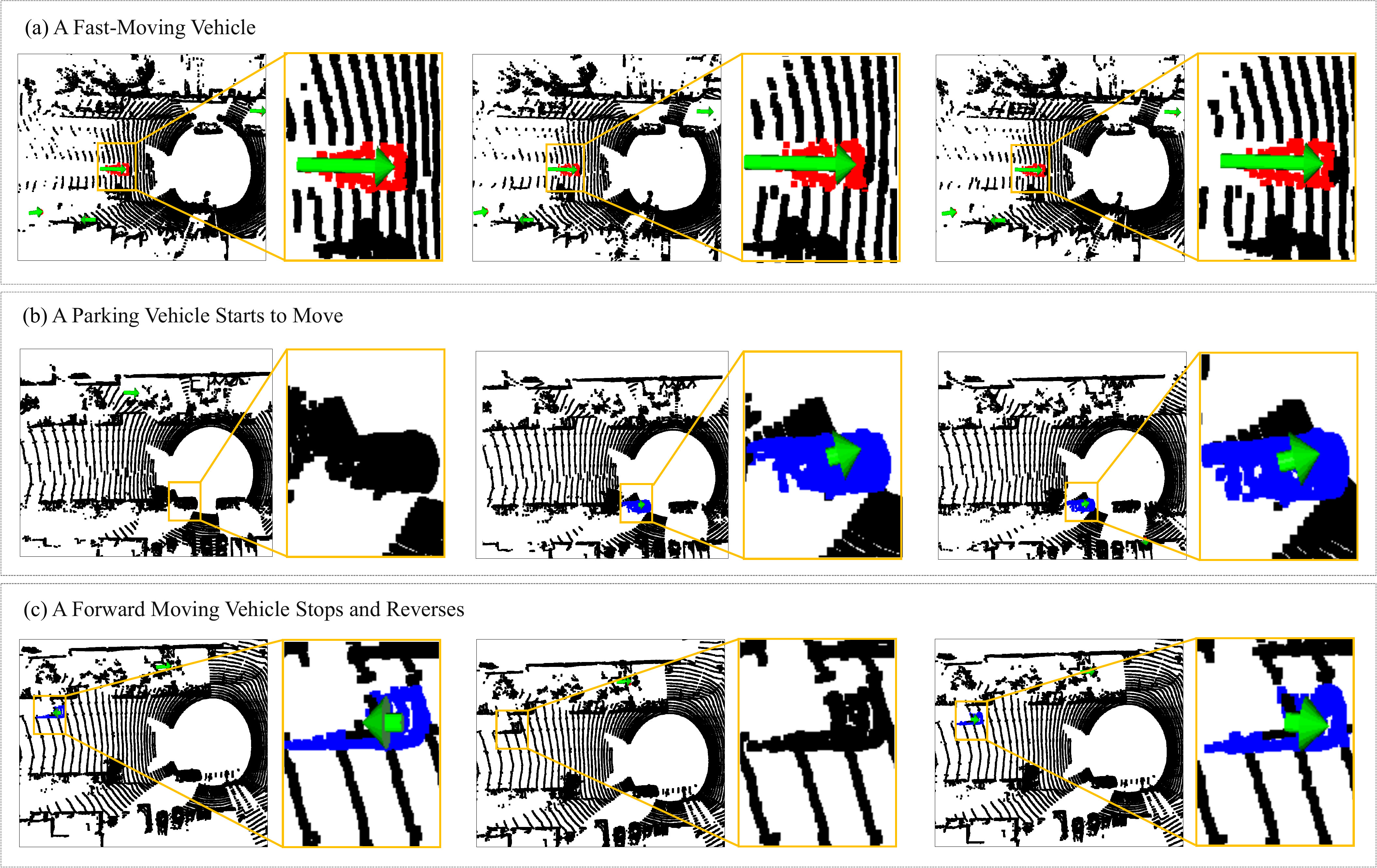}
\captionof{figure}{
\textbf{Illustration of Instantaneous Motion Perception.} We visualize motion of three objects from Waymo dataset~\cite{sun2020scalability}, each with three consecutive frames. Objects in fast and subtle motions are marked as {\color{red}red} and {\color{blue}blue}, respectively, with arrow length indicating motion magnitude. While standard motion detection handles general large motions such as (a), we focus on instantaneous perception of subtle motions that may indicate changes in driving behavior, for example (b) parking car starts to move, and (c) forward moving car stops and reverses. The visualized subtle motions (b)(c) are output from our framework. We also provide the video visualization in supplementary.
}
\label{cover_image}
\end{center}%
}]

\begin{abstract}
The perception of 3D motion of surrounding traffic participants is crucial for driving safety. While existing works primarily focus on general large motions, we contend that the instantaneous detection and quantification of subtle motions is equally important as they indicate the nuances in driving behavior that may be safety critical, such as behaviors near a stop sign of parking positions. We delve into this under-explored task, examining its unique challenges and developing our solution, accompanied by a carefully designed benchmark. Specifically, due to the lack of correspondences between consecutive frames of sparse Lidar point clouds, static objects might appear to be moving – the so-called swimming effect. This intertwines with the true object motion, thereby posing ambiguity in accurate estimation, especially for subtle motions. To address this, we propose to leverage local occupancy completion of object point clouds to densify the shape cue, and mitigate the impact of swimming artifacts. The occupancy completion is learned in an end-to-end fashion together with the detection of moving objects and the estimation of their motion, instantaneously as soon as objects start to move. Extensive experiments demonstrate superior performance compared to standard 3D motion estimation approaches, particularly highlighting our method's specialized treatment of subtle motions. 
\end{abstract}
    
\section{Introduction}

Human drivers pay special attention to surrounding moving objects to understand and predict their driving behavior, and react accordingly to avoid collisions. Similarly, intelligent autonomous systems must also navigate safely through traffic scenes, where preventing collisions with moving objects is considerably more complex than with static background scenes. This gives rise to several lines of computer vision research centered around motion, ranging from low-level tasks like 3D scene flow~\cite{Li_2023_ICCV}, to middle-level motion segmentation or detection~\cite{Wang_2022_CVPR}, and high-level perception on 3D object tracking~\cite{yin2021center}. 

However, these methods are designed to handle general 3D motion without considering the extent and context of the motion. In this paper, we would like to focus on an important subset of motion -- small subtle motions. Such motions are of significance as they often indicate changes in driving intention or behavior; for instance, as illustrated in \cref{cover_image}, parking vehicles start to move and cut into the driving lane, or vehicles in the driving lane start to reverse back for reverse parking. As a more general note, the instant capture of all nuanced changes happening in the scene is essential for situation awareness, especially in corner-scenario cases.
This however remains under-explored in computer vision, thus motivates our research in this paper, which aims to detect the presence of subtle motions as well as estimate their motion flow instantaneously.

While prominent motions from fast-moving objects are more feasible to detect and quantify due to a strong signal-to-noise ratio, recovering subtle motions with high accuracy presents its challenges. Specifically, the Lidar sensor captures only a sparse point set of the surrounding scene elements, and the pattern of points varies depending on the relative position between the Lidar and the scene. Consequently, there are typically no point correspondences across frames with a moving Lidar sensor, even for static scene elements. This further implies that static objects may appear to be moving, known as the swimming artifact~\cite{khurana2023point,chodosh2023re}. It intertwines with and hence obfuscates the true object motion, especially under small motions; as such, we empirically observe that a model trained for general motions does not perform as well with subtle motions.

To address this, our framework proposes to learn shape completion before performing motion detection and estimation. Taking sequential frames of Lidar point clouds within a short period as input, our method voxelizes the point clouds as occupancy grids, and leverages the accumulated Lidar points from nearby frames to generate a denser occupancy grid, which is then applied as supervision for occupancy completion. This effectively densifies and enhances the surface cue to mitigate the impact of the swimming artifact. To prevent the noises brought about by imperfect completion from affecting the downstream motion estimation task, we refrain from performing full object shape completion but rather do so locally only for the visible surface part, where the major motion signals lie.  
Our framework takes the point clouds of each object individually as input to our network, which is trained exclusively on the regime of small motion. We name our method \Ourstight, indicating subtle motion regressor. Since there is no standard training dataset and evaluation benchmark specific to subtle motions. We contribute one by extracting small motions from the large-scale Waymo dataset~\cite{sun2020scalability}, leveraging its existing annotations. We demonstrate the efficacy of the proposed method with the newly proposed benchmark.

\vspace{2mm}
\noindent In summary, our contributions include:
\begin{itemize}
  \item Introduce the task of detecting and estimating subtle motions for vehicles, with insights on their practical significance and challenges.
  \item Design a framework with occupancy completion to mitigate the swimming artifacts on small motion estimation.
  \item Translate our insights into favorable performance in a new evaluation benchmark tailored for small motions. 
\end{itemize}
\section{Related Work}

\begin{figure*}[t]
\vspace{-15pt}
\centering
\includegraphics[width=1\linewidth]{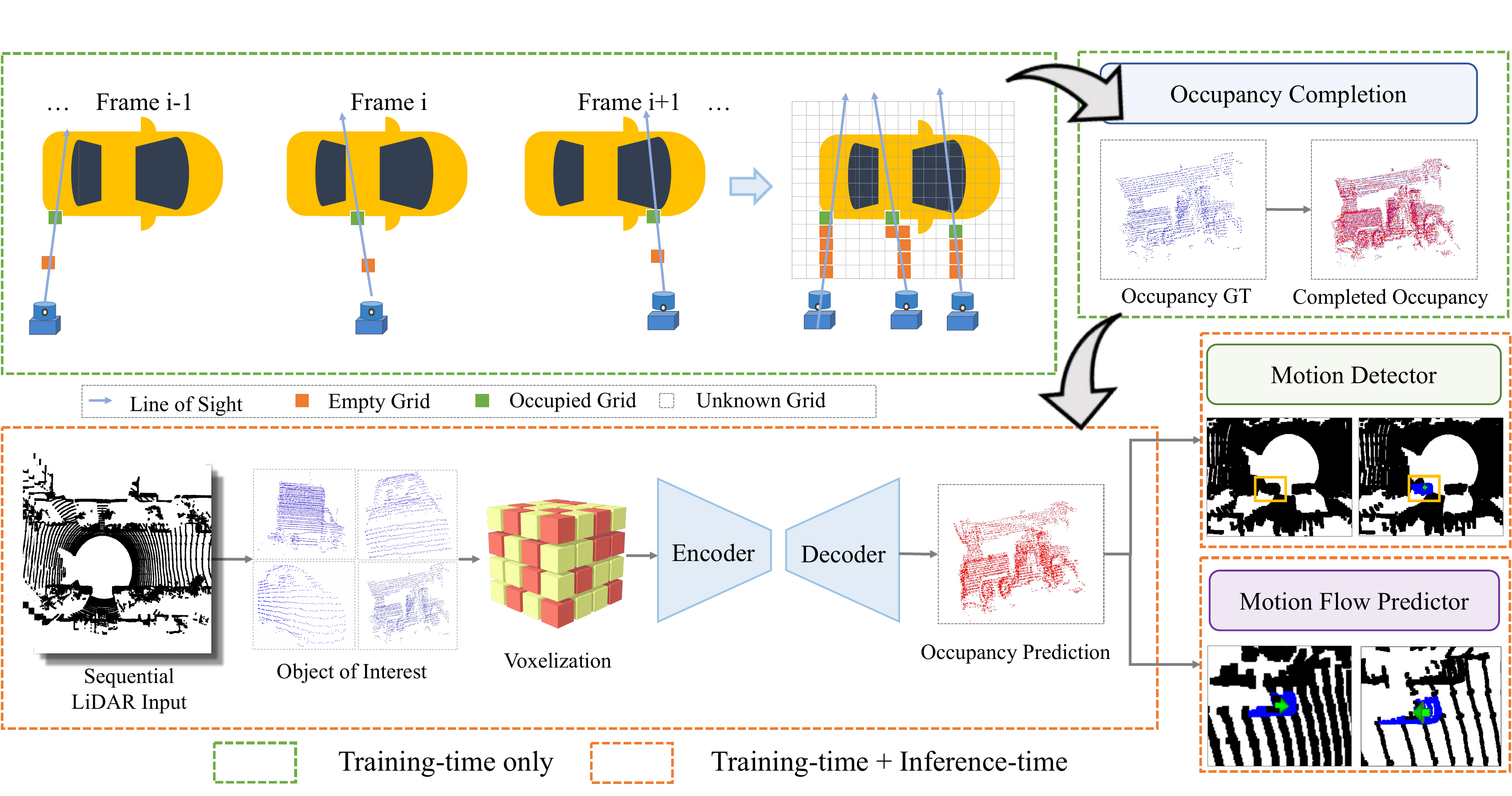}
\vspace{-15pt}
\caption{
\textbf{Overview of S'More.} Given a sequential LiDAR point cloud, we first identify objects of interest by filtering out background and objects with large motion. We then voxelize the point cloud for each object, extract features through an encoder-decoder network, and perform occupancy completion. The output is passed to a motion detector and a motion flow predictor for final detection and estimation. }
\label{fig:flowchart}
\end{figure*}
\noindent \textbf{3D Scene Flow.}
3D scene flow aims to estimate the motion field of each observed 3D point. It is an important tool for analyzing scene dynamics and has been extensively studied in computer vision~\cite{liu2019flownet3d, mittal2020just,li2021neural,zhang2022sequential,ding2023hidden,han2024proxedit,wen2024second,li2024steering}. While the scene flow of background points as the dominant rigid motion may be reliably estimated~\cite{chodosh2023re}, accurately estimating the motion flow for dynamic foreground objects remains a challenge. This leads to object-aware scene flow works \cite{behl2019pointflownet,li2021neural,vidanapathirana2023mbnsf,lee2020pillarflow}
that leverage rigidity prior of objects. However, the nearest neighbor-based approach for motion estimation like scene flow suffers from inherent ambiguity brought by the ``swimming" effect~\cite{khurana2023point} of Lidar point clouds, which is more severe with smaller motion magnitude. In this work, we develop insights towards addressing this issue.   


\noindent \textbf{Moving Object Detection.}  Moving object detection is an essential capability for autonomous vehicles and other areas, which results in many prior works, such as ~\cite{Wang_2022_CVPR, mersch2023building,mersch2022receding,huang2022dynamic,sun2022efficient,gao2023training,filatov2020any,liu2022transfusion,chang2022deeprecon}. However, they detect motions at a coarse level for general large motions. SemanticKITTI~\cite{behley2019semantickitti}, a commonly used dataset in this field, labels moving object in a coarse sequence level instead of in an instantaneous manner. 
In contrast, our approach offers special treatment for detecting small motions instantaneously. Another possible way for moving object detection is through 3D detection~\cite{liu2023bevfusion,min2023neurocs,he2023dealing,gao2022data,zhangli2022region,liu2021refined,liu2021label,martin2023deep} and tracking~\cite{yin2021center,huang2023delving}. However, we found empirically that such methods stumble in identifying small motions due to imperfect object localization.
Lastly, we also note a concurrent work, M-detector~\cite{wu2024moving}, that instantly detects point-level moving events based on occlusion principles. 

\noindent \textbf{Occupancy Prediction.}
Occupancy is an effective 3D scene representation that has wide applications in autonomous driving. Argo et al.~\cite{agro2023implicit} and Reza et al.~\cite{mahjourian2022occupancy} apply the occupancy flow field for perception and motion forecasting. \cite{khurana2023point} performs 4D occupancy forecasting supervised by point cloud forecasting. ALSO~\cite{boulch2023also} utilizes occupancy completion as a tool for self-supervised feature learning for Lidar point clouds. Another line of work~\cite{tong2023scene,wei2023surroundocc,wang2023openoccupancy} learns occupancy prediction from monocular cameras. In this work, we present the first known attempt to use occupancy completion to facilitate the estimation of small motions from LiDAR point clouds.

\noindent \textbf{Subtle Motions in General.} The field of computer vision has shown a long history of interest in small motions in a broader context. Differential structure-from-motion~\cite{ma2000linear,zhuang2017rolling} aims to recover instantaneous camera motion from optical flow. Several works~\cite{yu20143d,ha2016high,chugunov2022implicit,liu2024lepard,liu2023deformer,liu2023deep} utilize accidental camera motion to perform 3D reconstruction. Another line of research~\cite{liu2005motion,feng20233d} targets magnifying invisible subtle motions in videos. Our work focuses on perceiving the subtle motions of surrounding vehicles, a capability critical to the safety of autonomous systems.
\section{Method}

\subsection{Problem Definition and Challenges}

\begin{figure}[t]
\centering
\includegraphics[width=1.0\linewidth, trim = 5mm 2mm 5mm 2mm, clip]{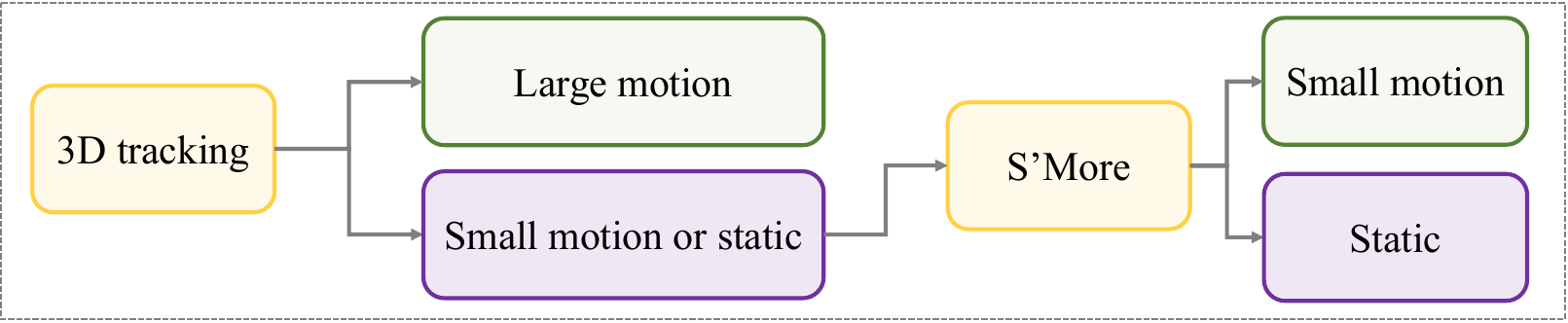}
\caption{
\textbf{Integration of S'More with 3D tracking systems.}}
\label{fig:withtracking}
\vspace{-5pt}
\end{figure}

Our goal is to identify moving objects and estimate their motion using sequential point clouds; we focus in this paper on vehicles while leaving the human category for future work. Unlike existing works, we concentrate on small motion for instantaneous detection as objects begin to move. As a preprocessing step, our framework filters out fast-moving objects, thereby targeting static and slow-moving ones. \cref{fig:withtracking} illustrates a practical use case of such setting in a 3D tracking system.
Further, we make a practical assumption that the ego-vehicle's motion can be reliably estimated by ICP, possibly aided by GPS/INS, as validated by recent studies~\cite{chodosh2023re}. This allows for the exclusion of ego-motion from the observed object motion, resulting in a noisy observation of the true object motion \textit{w.r.t.} the world coordinate. Our method is object-centric, processing point clouds from five consecutive frames ($F_t, t{=}1,...,T$) to classify objects as static or moving. For moving objects, we estimate the motion flow from $F_1$ to $F_T$ for each point in $F_1$, setting $T{=}5$ as per~\cite{huang2022dynamic}.

\begin{figure}[t]
\centering
\includegraphics[width=1\linewidth]{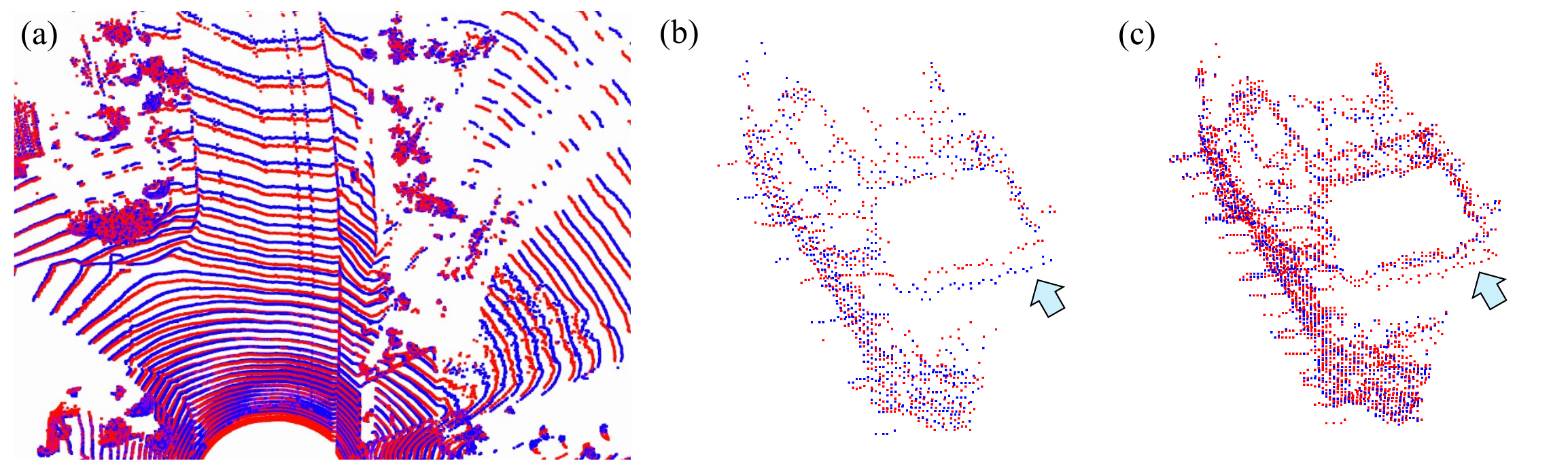}
\caption{
\textbf{Illustration of Swimming Effect} on ground (a) and static object (b), and (c) our ground truth occupancy completion for (b). Bule and red points indicate points from two frames.}
\label{fig:methodswimmingeffect}
\end{figure}

\noindent \textbf{Swimming Effect.} The detection and estimation of small motions present its challenges, primarily arising from the sparse nature of Lidar point clouds. Remarkably, the spatial distribution of captured points closely depends on the relative position between the Lidar sensor and surrounding scene elements. Hence, as the Lidar sensor moves along with the ego-vehicle, there are typically no exact point correspondences across frames, and even static scene elements may appear to be moving. This effect manifests itself on both background scene and foreground objects, as illustrated in \cref{fig:methodswimmingeffect}~(a)(b). In particular, the ground points visually appear to be swimming across frames, hence termed as \textit{swimming effect}~\cite{khurana2023point}. This effect poses challenges, especially to characterizing subtle motions as one would need to distinguish the true object motion from this effect.
We note that the sparse nature of Lidar points distinguishes the problem from optical flow~\cite{dosovitskiy2015flownet}, 
where dense correspondences exist and small motions simplify the flow estimation through brightness consistency assumption.

\subsection{Our Framework}
\cref{fig:flowchart} illustrates the overview of \Ourstight. Given five consecutive Lidar frames,  
we filter out fast-moving objects and for each remaining ones, 
our network estimates the small motion or the lack thereof. We start by voxelizing the input point clouds, followed by feature extraction using an encoder-decoder network, and then perform occupancy completion, the output of which is passed to the motion segmentation head and the instantaneous flow estimation head.

\subsubsection{Occupancy Completion}
\noindent \textbf{Input Voxelization.} First, following state-of-the-art 3D detection frameworks~\cite{yin2021center,liu2023bevfusion}, we voxelize the point cloud $\bf{X}_t \in R^{N\times3}$ at each frame as a binary grid of size $[W_x, W_y, W_z]$, with the voxels containing Lidar points filled with 1, and 0 elsewhere. This grid may be viewed as an incomplete occupancy grid, in that it indicates part of the visible object surface captured by Lidar at a single timestamp. We stack consecutive frames to form a spatial-temporal grid of size $[T, W_x, W_y, W_z]$.

\noindent \textbf{Local Occupancy Completion.} Recall that the sparse nature of Lidar point clouds poses challenges to accurate small motion estimation, due to the actual subtle motion intertwined with the swimming effect. To mitigate this issue, our framework first learns occupancy completion that effectively densifies the object surface, to offer stronger cues for subsequent networks to reason correspondence, and hence the motion between frames. 

Before proceeding, one should be mindful of the potential trade-off brought by this step -- the estimated occupancy completion may well be imperfect, introducing extra noise to the system. This may be harmful to the final motion segmentation and estimation if the noises reach a certain level, hence defeating the purpose of occupancy completion. Therefore, while the standard shape completion problem (\eg~\cite{park2019deepsdf}) is tasked to recover the entire object shape from a single-frame input, it is an overly complicated and unnecessary task in our case, besides the infeasibility of getting the ground truth in real driving scenes. Instead, since the $T$ Lidar frames collectively observe only a local part of the object within an instantaneous timeframe, we target local occupancy completion at these observed regions, while refraining from hallucinating areas invisible to all $T$ Lidar frames. This way, we enhance the critical signal essential for motion characterization while minimizing the extra noises from imperfect completions.

\noindent \textbf{Supervision for Occupancy.} We densify the local occupancy grid by leveraging nearby frames, as shown in Fig.~\ref{fig:flowchart}. Specifically, for each frame $F_t, t\in [1,...,T]$ in the input window, we warp all Lidar points from the rest $T-1$ frames to $F_t$, using the ground truth object motion (recall that ego-motion has been factored out), and then mark the corresponding voxel as occupied, \ie 1. In addition, we mark the points along the line-of-sight as empty \ie 0. All other voxels are deemed as unknown.
We apply a fast voxel traversal algorithm~\cite{amanatides1987fast} to implement this step, similarly as in ~\cite{khurana2023point}, with an example illustrated in \cref{fig:methodswimmingeffect}(b)(c).
Note we only use the Lidar frames inside the input window to generate the target occupancy grid, to simplify the task. 
By learning occupancy, the network is explicitly enforced to learn the notion of dense shapes in an end-to-end manner, thereby facilitating the task of motion detection and estimation. 

\subsubsection{Network Architecture and Losses}
\noindent \textbf{Network Architecture.} We apply an encoder-decoder for occupancy grid prediction, which is passed to another encoder-decoder for motion detector and motion flow predictor. The motion detector classifies input objects as static or moving, while the flow estimator regresses a motion vector for each occupied voxel in the grid. We then extract the motion flow for each raw input point as the predicted flow in the voxel that point resides in.
Note that we do not enforce rigidity constraints on the flow field, maintaining the method's generality, though we do evaluate the setting with rigidity prior as well.  We utilize the encoder-decoder structure as in~\cite{zeng2019end}, consisting of simple convolutional layers with skip connection; we follow~\cite{khurana2023point} to treat the height and temporal dimension as the channel dimension, which allows to use 2D convolutional layers for efficiency; see supplementary for details. Our network processes each object separately, but remains efficient and runs at 27 fps for a scene consisting of 30 objects of interest.

\noindent \textbf{Overall Losses.} The overall loss function of our model is a weighted combination of five terms:
\begin{equation}
    \mathcal{L} = \lambda _ \text{occ} \mathcal{L_\text{occ}} + \lambda _ \text{mot} \mathcal{L_\text{mot}} + \lambda _ \text{epe} \mathcal{L_\text{epe}} + \lambda _ \text{rel} \mathcal{L_\text{rel}} + \lambda _ \text{ang} \mathcal{L_\text{ang}}.
\end{equation} 
Specifically, we apply a binary cross-entropy (BCE) loss $\mathcal{L_\text{occ}}$ for the occupancy grid prediction, a BCE loss $\mathcal{L_\text{mot}}$ on static/moving object classification, a $L_1$ loss $\mathcal{L_\text{epe}}$ and a scale-aware $\mathcal{L_\text{rel}}$ loss on motion flow prediction for moving objects. Additionally, since the motion direction carries important information about driving intention such as reversing or left/right turning, we add an angular loss $\mathcal{L_\text{ang}}$ for the motion flow. We denote the set of occupied and empty  voxels as $\phi_o$ and $\phi_e$, respectively.   

\noindent \textbf{Occupancy Loss} is written as  
\begin{equation}
\mathcal{L}_{\text{occ}} = \mathbb{E}_{v \in \{\phi_o,\phi_e\}}\left[ {\hat O}_v \log({O}_v) + (1 - {\hat O}_v) \log(1 - {O}_v) \right], 
\end{equation}
where $O_v$ and $\hat O_v$ indicate the predicted and ground truth occupancy at voxel $v$.

\noindent \textbf{Flow Prediction Losses.}
For each voxel $v$, we define the ground truth flow (denoted as $\hat{\mathbf{f}}_v$) as the average of the ground truth flow associated with the points falling into that voxel. 
The relative flow loss $\mathcal{L}_{\text{rel}}$ is written as 
\begin{equation}
\mathcal{L}_{\text{rel}} = \mathbb{E}_{v \in \phi_o} \frac{\| \hat{\mathbf{f}}_v - \mathbf{f}_v \|_2}{\| \hat{\mathbf{f}}_v \|_2 + \varepsilon},    
\end{equation}
where $\hat{\mathbf{f}}_v$ indicates predicted flow at voxel $v$ and $\varepsilon$ is a small constant.
Note this loss is inverse scaled by the flow magnitude to emphasize learning on small motions.
And the augular loss is written as
\begin{equation}
\mathcal{L}_{\text{ang}} = \mathbb{E}_{v \in \phi_o} \text{acos}\left( \frac{\langle {\mathbf{f}}_v, \hat{\mathbf{f}}_v \rangle}{\| {\mathbf{f}}_v \|_2 \cdot \| \hat{\mathbf{f}}_v \|_2 + \varepsilon} \right)    
\end{equation}
where $\langle \cdot, \cdot \rangle$ denotes the dot product between the vectors.

\section{Experiments}
\label{experiments}

\begin{table}[t]
\centering
\caption{\textbf{Quantitative Comparisons} with ICP, Point-to-Plane ICP, Generalized ICP, CenterPoint and FastNSF.  
}
\renewcommand\tabcolsep{7pt}
\resizebox{1\linewidth}{!}{
\begin{tabular}{lccc}
\toprule
\multirow{1}{*}{}
& EPE ($\downarrow$) & Angle Error ($\downarrow$) & F1 Score ($\uparrow$) 
\\ 
\midrule
FastNSF~\cite{Li_2023_ICCV} & 0.1189 & 0.5592 & 0.6180 \\
ICP~\cite{besl1992method}
& 0.0554 &  0.4499 & 0.7456\\
Point-to-Plane ICP~\cite{rusinkiewicz2001efficient} & 0.2263 & 0.4379 & 0.7856\\
Generalized ICP~\cite{segal2009generalized} & 0.1117 & 0.4170 & 0.7693 \\
CenterPoint~\cite{yin2021center} & 0.0927 & 0.5622 & 0.7270 \\
S'More    
& \textbf{0.0437}  & \textbf{0.3189} &  \textbf{0.8323} \\
\bottomrule
\end{tabular}} 
\label{tab_comp}
\end{table}

\begin{figure*}[t]
\centering
\includegraphics[width=0.96\linewidth]{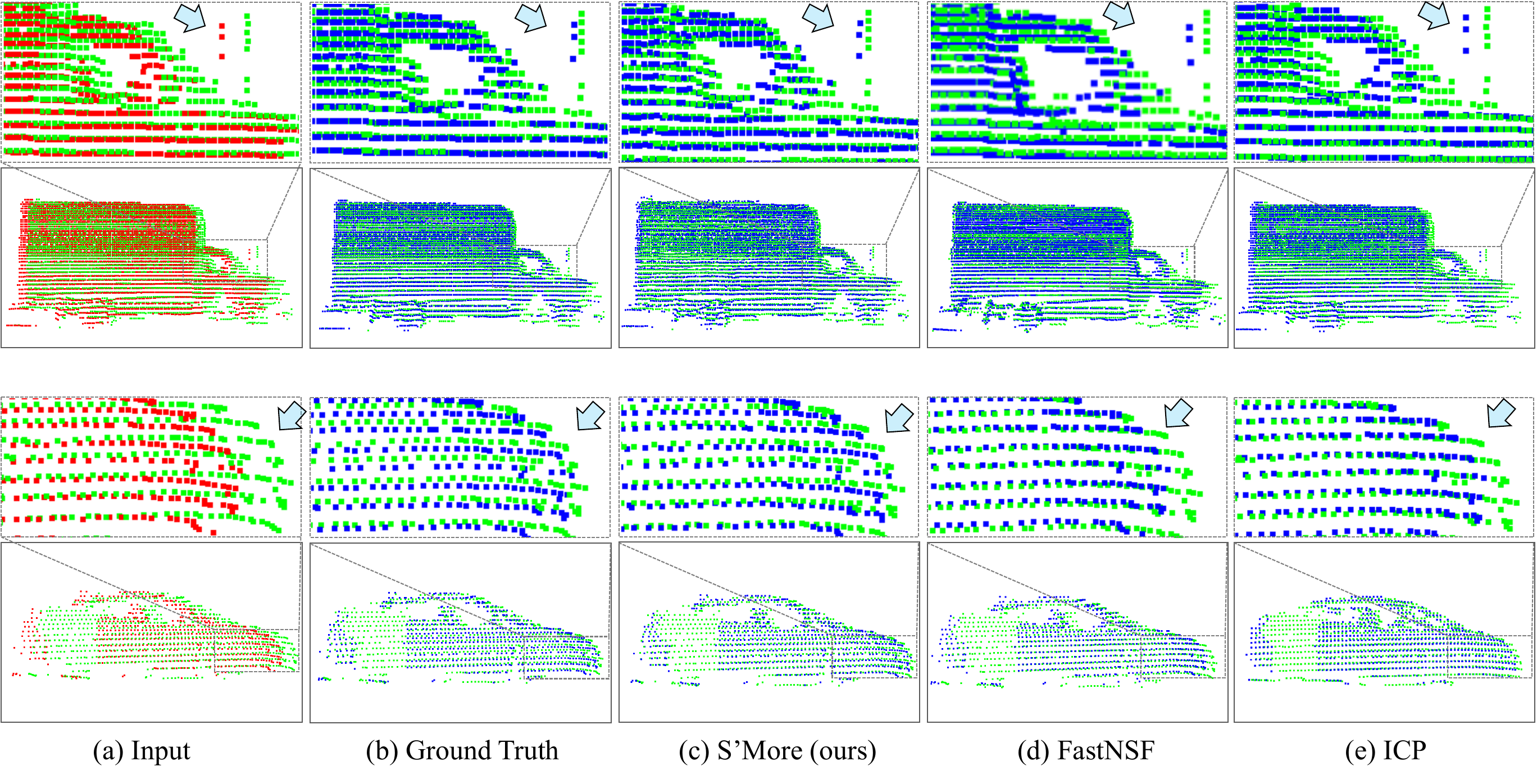}
\vspace{-4mm}
\caption{
\textbf{Qualitative Comparison.} We exhibit point cloud registration results for two point cloud sets: the first frame (in {\color{red}red}) and the last frame (in {\color{green}green}). The results are shown using (b) ground truth motion, and estimated motions by (c) \Ourstight (ours), (d) FastNSF, and (e) ICP. The {\color{blue}blue} points indicate resultant positions after adding flow to the red points, which should ideally align with the green points.}
\label{vis_comp}
\end{figure*}

\begin{figure*}[t]
\centering
\includegraphics[width=1\linewidth]{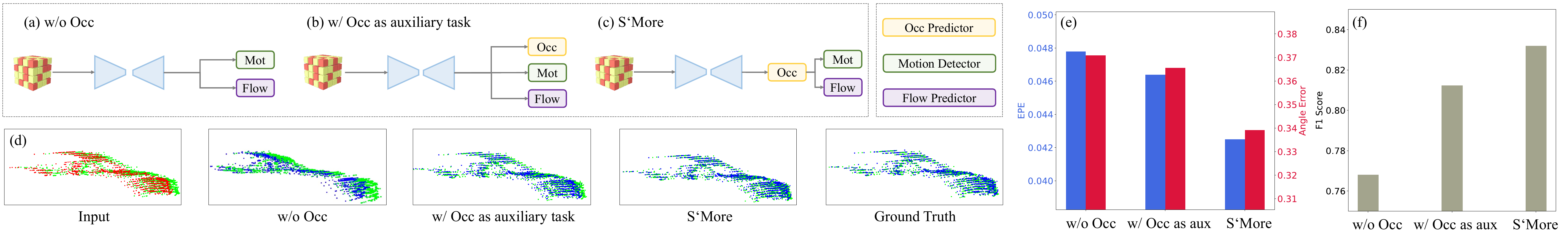}
\caption{
\textbf{Ablation study on Occupancy Completion.} (a)(b)(c) illustrate structure of having the occupancy module removed or as an auxiliary task. (d)(e)(f) show the qualitative and quantitative comparisons; the visualization in (d) follows the same protocol in \cref{vis_comp}.}
\label{chart_deep_aux}
\vspace{-4mm}
\end{figure*}

\begin{figure}[t]
\centering
\includegraphics[width=1\linewidth]{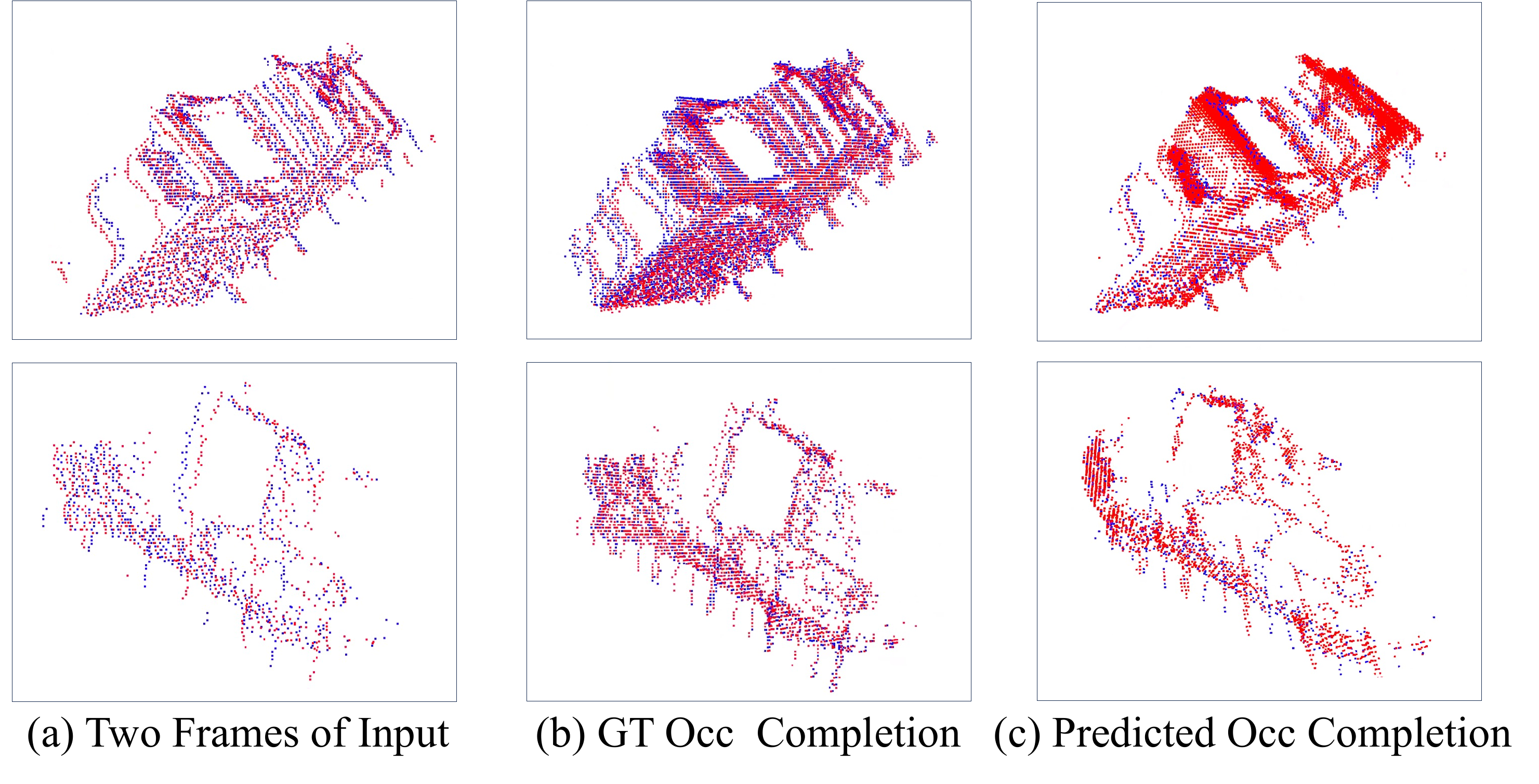}
\caption{
\textbf{Example Results of Occupancy Completion.} The blue and red points represent different frames in the input.}
\label{fig:vis_occ_completion}
\vspace{-5pt}
\end{figure}

\begin{figure*}[t]
\centering
\includegraphics[width=1\linewidth]{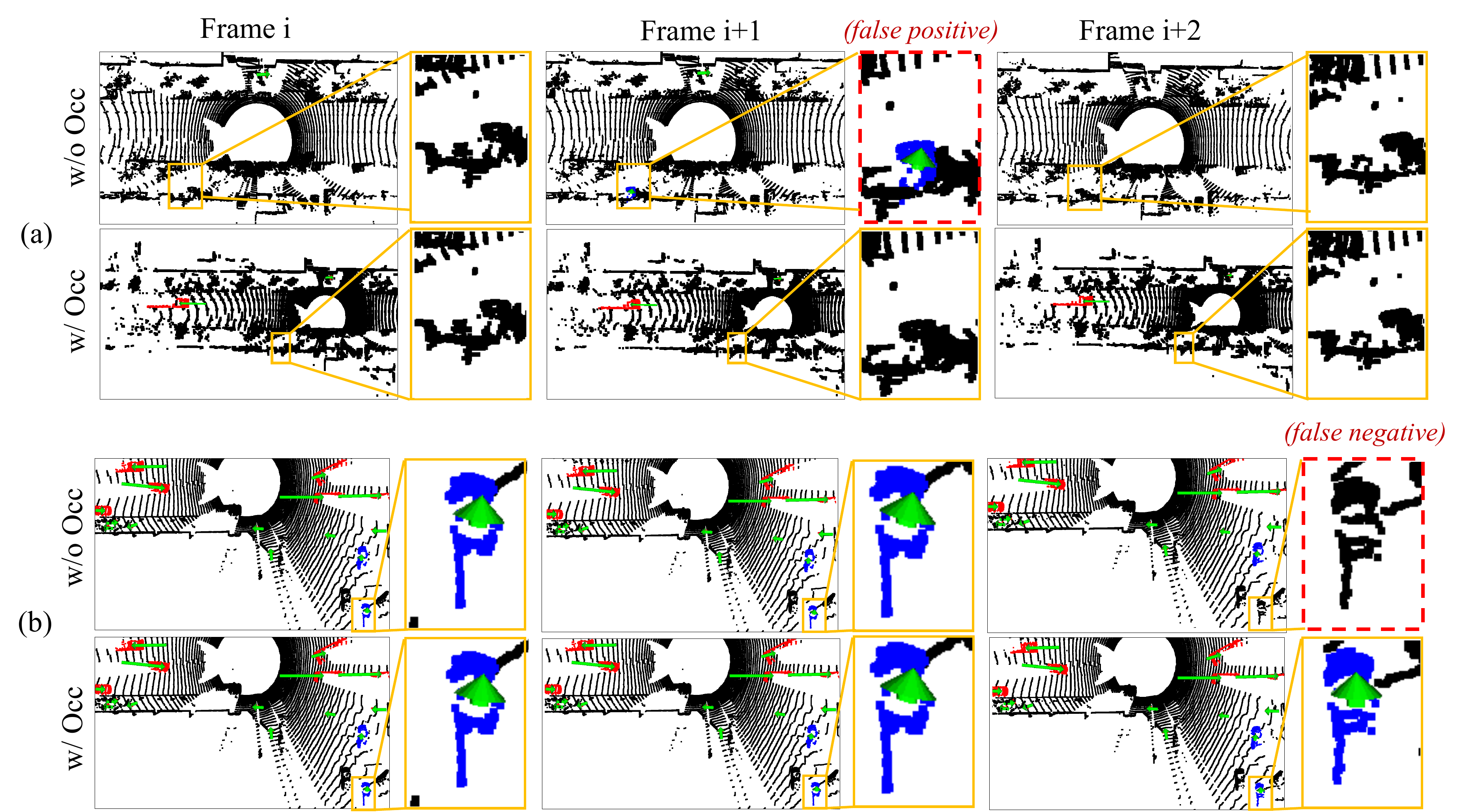}
\caption{
\textbf{Visualization of false positive/negative samples in the absence of occupancy completion.} Each row shows three consecutive frames, one per column. The motion of objects is marked with {\color{blue}blue} for detected subtle motions and {\color{red}red} for GT large motions, with the arrow length representing the motion's magnitude. In (a), a false positive occurs where a stationary vehicle is mistakenly marked as moving. In (b), a moving vehicle is incorrectly identified as static, a false negative. 
Both of them are rectified with the occupancy completion; note the ground truth is same as the correct detection here hence is not visualized. }

\label{vis_fp_v}
\end{figure*}

\subsection{Evaluation of \Ours}
\noindent \textbf{Evaluation Benchmark.} In the absence of existing benchmark dedicated to subtle motions, we curate one such dataset from the Waymo open dataset~\cite{sun2020scalability}, where each sequence provides Lidar frames at 10Hz for about 20s. We collect the point clouds for each object from every five consecutive frames (0.5s) denoted as $F_i, i{=}[1,...,5]$. To generate ground truth motion status, we follow~\cite{huang2022dynamic} to derive the spatial transformation from the 3D boxes annotations, and hence compute the scene flow $\mathbf{f}_i$ from $F_1$ to $F_5$ for every point $\mathbf{x}_i$ in $F_1$. To concentrate on small motions, we deem the data sample valid only when the minimal flow magnitude $f_\text{min}{=}\min_{\mathbf{x}_i \in F_1} ||\mathbf{f}_i||$ is less than 0.2m. Further, we label the object as static if $f_\text{min}{<}f_\text{thre}$. We set $f_\text{thre}{=}0.05$m but also evaluate under other settings shortly. This way, we collect about 140k and 9k samples for training and test sets, respectively. More details are in the supplementary.  

\vspace{0.1cm}
\noindent \textbf{Evaluation Metrics.}
We apply the standard F1 score to measure the accuracy of static/moving object classification. We apply end-point error (EPE) and the angular error to measure the object motion flow error. 

\vspace{0.1cm}
\noindent \textbf{Baselines.} In the absence of existing detection methods dedicated to small motions, we mainly compare with: (i) the classical Iterative Closest Point (ICP)~\cite{besl1992method}, which remains competitive~\cite{chodosh2023re} for motion flow task; (ii) the point-to-plane ICP~\cite{rusinkiewicz2001efficient} and the generalized ICP~\cite{segal2009generalized} implemented in Open3D~\cite{Zhou2018}; (iii) the leading scene flow method FastNSF~\cite{Li_2023_ICCV}; (iv) the detection and tracking based method CenterPoint~\cite{yin2021center}, where we use ground-truth tracking by associating detected objects with the ground truth, and the motion flows are derived from boxes transformation. For all methods we use their output motion flows to detect moving objects, according to the aforementioned criterion.

\vspace{0.1cm}
\noindent \textbf{Comparison.} \cref{tab_comp} shows quantitative evaluation results, indicating the significantly superior performance of our model compared to the baselines. We note that the object-tracking method CenterPoint gives decent accuracy but lags behind S'More, likely because their imperfect object localization causes ambiguity in distinguishing small motions from static ones.
In Fig.~\ref{vis_comp}, we provide a visualization comparison with ICP and FastNSF. The input comprises two sets of point clouds: the first frame (in {\color{red}red}) and the last frame (in {\color{green}green}). We visualize the flow accuracy through alignment -- we shift the red points with the flow and the resultant points (marked as {\color{blue}blue}) should ideally align well with the green points if the flows are correct.
Our model demonstrates superior alignment accuracy, especially at the subtle level of local registration,
attributable to its advanced motion estimation capabilities. More examples are provided in the supplementary material.
\subsection{Ablation study on Occupancy Completion}

\begin{table*}[t]
\centering
\caption{\textbf{Ablation Study of Losses.} This study observes a correlation between decreasing flow thresholds $f_\text{thre}$ and the degradation of model performance. Notably, our model achieves the best average performance over five thresholds in terms of all metrics. }
\vspace{-5pt}
\resizebox{1\linewidth}{!}{
\begin{tabular}{lcccccccccccccccccc}
\toprule
\multirow{2}{*}{}
& \multicolumn{6}{c}{End-Point-Error (EPE) ($\downarrow$) } & \multicolumn{6}{c}{Angle Error ($\downarrow$) } & \multicolumn{6}{c}{F1 Score ($\uparrow$)} \\ 
\cmidrule(lr){2-7}\cmidrule(lr){8-13}\cmidrule(lr){14-19}
$f_\text{thre}$      
& 0.05 & 0.04 & 0.03 & 0.02 & 0.01  & Avg & 0.05 & 0.04 & 0.03 & 0.02 & 0.01  & Avg & 0.05 & 0.04 & 0.03 & 0.02 & 0.01  & Avg \\
\midrule
w/o $\mathcal{L_\text{occ}}$
& 0.0492 & 0.0443 & 0.0440 & 0.0489 & 0.0439 & 0.0461 & 0.3489 & 0.3455 & 0.3828 & 0.4219 & 0.4405 & 0.3879 & 0.8256 & 0.8202 & 0.8025 & 0.7575 & 0.7603 & 0.7932\\

w/o $\mathcal{L_\text{angle}}$
& 0.0458 & 0.0441 & 0.0452 & 0.0444 & 0.0454 & 0.0450 & 0.3469 & 0.3654 & 0.4052 & 0.4295 & 0.4860 & 0.4066 & 0.8344 & \textbf{0.8338} & \textbf{0.8111} & 0.7704 & 0.7841 & 0.8068 \\
w/o $\mathcal{L_\text{rel}}$  
& 0.0446 & 0.0435 & 0.0439 & 0.0437 & 0.0430 & 0.0437 & \textbf{0.3179} & 0.3395 & \textbf{0.3780} & 0.4156  & \textbf{0.4291} & 0.3760 & \textbf{0.8378} & 0.8307 & 0.7937 & 0.7907 & 0.7847 & 0.8075 \\

S'More    
& \textbf{0.0437} & \textbf{0.0425} & \textbf{0.0436} & \textbf{0.0428} & \textbf{0.0421}& \textbf{0.0429}  & 0.3189 & \textbf{0.3392} & 0.3834 &  \textbf{0.4053}  & 0.4324 & \textbf{0.3758} & 0.8323 & 0.8320 & 0.8003 & \textbf{0.7997} & \textbf{0.7857} & \textbf{0.8100}\\
\bottomrule
\end{tabular}} 
\label{tab_loss}
\vspace{-5pt}
\end{table*}



To investigate the impact of occupancy completion, we \textit{remove} this module from \Ourstight, with structure shown in \cref{chart_deep_aux}(a) in relation to \Ours in \cref{chart_deep_aux}(c). Further, we also evaluate the setting with the occupancy completion as just an auxiliary trained in parallel with the motion detector and flow predictor, as shown in \cref{chart_deep_aux}(b).
We report the accuracy in Fig.~\ref{chart_deep_aux}(e)(f), which indicates the significant impact of occupancy completion towards good performance. We attribute this to its role in effectively densifying object surfaces. In Fig.~\ref{chart_deep_aux}(d) we provide visualization of point cloud registration to evaluate the estimated motion, further supporting the efficacy of occupancy completion. The qualitative results of the occupancy completion itself are demonstrated in \cref{fig:vis_occ_completion}.


\vspace{0.1cm}
\noindent \textbf{False Positive/Negative.} This analysis explores how incorporating occupancy impacts the reduction of false positives and negatives. In Fig.~\ref{vis_fp_v}(a), we display three sequential frames that highlight how the model, without occupancy, erroneously detects a stationary vehicle as moving, marked by a red box. Conversely, when employing dense occupancy estimation, the model correctly identifies the stationary nature of the vehicle, reducing the false positives. Similarly, in Fig.~\ref{vis_fp_v}(b), we show that without occupancy, the model fails to detect a moving vehicle as static. This is rectified using the model with occupancy. This comparison underscores the effectiveness of using occupancy data in enhancing motion detection accuracy in our model. An interesting observation we found is that the occurrence frequency of false positives is much higher than false negatives. We attribute this to the swimming effect, where static objects appear to be moving, especially under subtle motion conditions.

\begin{figure}[t]
\centering
\includegraphics[width=1\linewidth]{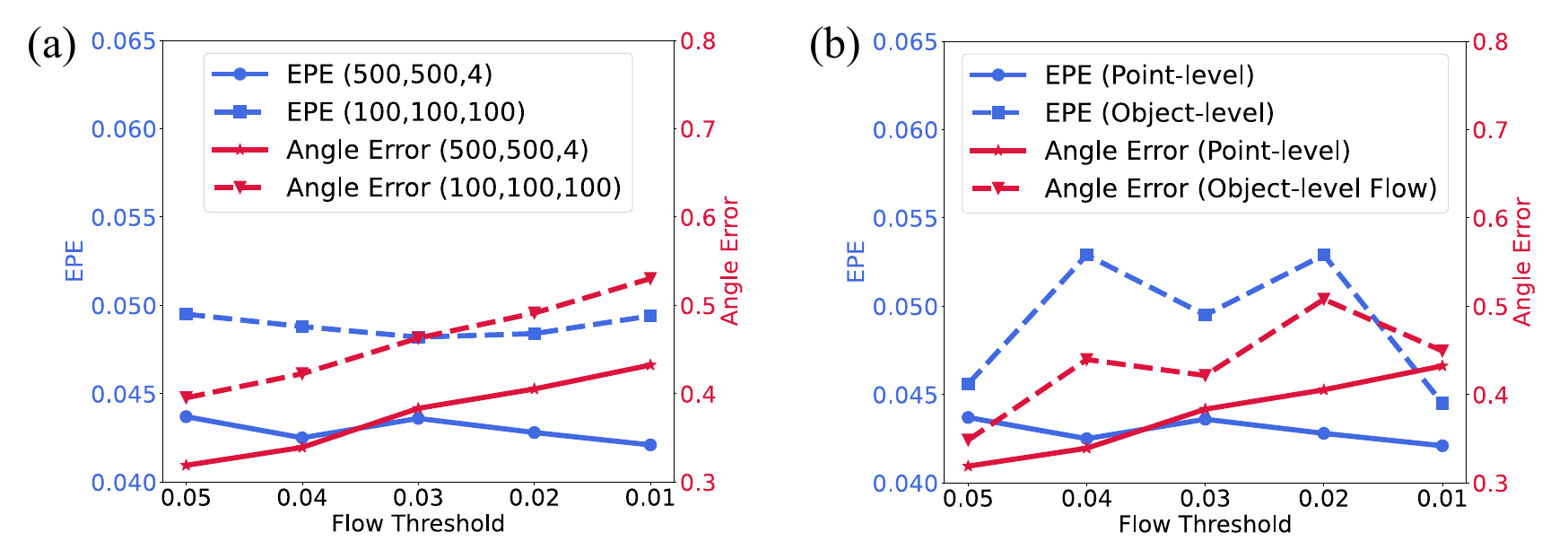}
\caption{
\textbf{Performance Analysis} of (a) occupancy with different grid sizes and (b) point-/object-level flow estimation. We evaluate under different value of $f_\text{thre}$ for comprehensiveness. }
\label{chart_grid_RT}
\vspace{-9pt}
\end{figure}

\subsection{Analysis on Large Motion}

\noindent \textbf{Performance on large motion.} 
While our focus is on small motions, it would be more complete to study its performance on large motions as well. It is of particular interest to compare against 3D object tracking based method given the practical system illustrated in \cref{fig:withtracking}. To this end, we train another model of \Ours including large-motion data and evaluate only on the large motion ($f_\text{min}>$0.2m) regime. 
As shown in \cref{tab:largemotion_tracking}, both S'More and CenterPoint achieve nearly perfect detection accuracy (F1$\to$1.0), as expected due to the large signal-to-noise ratio. This perfectness supports our focus on the small motion regime for enhancing a practical system. We also note that being a detection-tracking method, CenterPoint yields more precise flow estimation, as its accuracy largely depends on 3D box localization instead of motion.    

\begin{table}[t]
\centering
\caption{\textbf{Quantitative evaluation on large motions}.}  
\vspace{-1mm}
\renewcommand\tabcolsep{6pt}
\resizebox{1\linewidth}{!}{
 \begin{tabular}{lccc}
\toprule
\multirow{1}{*}{}
& EPE ($\downarrow$) & Angle Error ($\downarrow$) & F1 Score ($\uparrow$)   
\\ 
\midrule
CenterPoint~\cite{yin2021center}& \textbf{0.1155} & \textbf{0.0247} & \textbf{0.9998} \\
S'More & 0.2790 & 0.0486 & 0.9847\\ 
\bottomrule
\end{tabular}} 
\label{tab:largemotion_tracking}
\end{table}

\noindent \textbf{Benefit of using small-motion-specific dataset.} Recall that we have filtered out the large-motion data during the dataset curation, \ie they are not in training data. Our experience is that excluding large motions in training data improves the model performance on the small motion regime. We report the accuracy in \cref{tab:largemotion_data} to quantify this effect. The benefit may be explained by the unique swimming artifact, which mandates a small-motion-specific dataset.      

\begin{table}[t]
\centering
\caption{\textbf{Benefits of having small-motion-specific training data}.}  
\renewcommand\tabcolsep{4pt}
\resizebox{1\linewidth}{!}{
\begin{tabular}{lcccc}
\toprule
\multirow{1}{*}{Training dataset}
 & EPE ($\downarrow$) & Angle Error ($\downarrow$) & F1 Score ($\uparrow$) 
\\ 
\midrule
Small + large motion  & 0.0979 & 0.5550 & 0.1414 \\
Small motion only & \textbf{0.0437}  & \textbf{0.3189} &  \textbf{0.8323} \\
\bottomrule
\end{tabular}} 
\label{tab:largemotion_data}
\end{table}

\subsection{Evaluation in terms of Latency}
With a focus on instantaneous detection, a time-sensitive task, it is helpful to also evaluate with a time-related metric. Our original task is to detect objects moving more than $f_\text{thre}{=}0.05$m in a 0.5s latency. Here, we increase $f_\text{thre}$ to target at larger motion, which effectively allows proportionally increased latency if assuming constant velocity, hence decreasing the requirement on the latency. We report in \cref{tab:latency} the detection accuracy (F1) across different latencies, indicating the consistently superior performance from \Ourstight.

\subsection{Important Design Choices}
\vspace{0.1cm}
\noindent \textbf{Grid Size.} We study the impact of occupancy grid size and find it important in our design. We compare the performance of two grid sizes: a balanced $100 \times 100 \times 100$ grid and an alternative $500 \times 500 \times 4$ grid, where the latter significantly reduces the resolution along $z$-axis. The results in \cref{chart_grid_RT}(a), reveal a notable performance degradation (see the two dashed lines are consistently lower than the solid lines) when the $z$-axis resolution is reduced, despite the increased axial resolution to $500 \times 500$. This may also result in detection ambiguity which stems from the model's reduced capacity to discern subtle vertical variations, and lead to less reliable vehicle localization and motion detection. 



\vspace{0.1cm}
\noindent \textbf{Point-/Object-level Instance Flow Estimation.} We study the behavior of the point-level and object-level training strategies for motion flow estimation. The point-level approach predicts a separate flow for each point, whereas the object-level strategy regresses a single rigid transformation for the entire object to calculate motion flow. The results detailed in Fig.~\ref{chart_grid_RT}(b) in terms of EPE and Angle Error across various flow thresholds, reveal that point-level instance flow estimation (indicated by two solid lines) consistently outperforms the object-level approach (indicated by two dashed lines). Notably, point-level estimation maintains stable performance even at very low thresholds. In contrast, object-level flow prediction exhibits significant fluctuations in performance across different thresholds. 

\begin{table}[t]
\centering
\caption{\textbf{Evaluation of detection (F1 score) in terms of latency.}}
\renewcommand\tabcolsep{5pt}
\resizebox{1\linewidth}{!}{
\begin{tabular}{lcccccccc}
\toprule
\multirow{1}{*}{Latency}
 & 0.5s & 0.7s & 0.9s & 1.1s & 1.3s 
\\ 
\midrule
ICP~\cite{besl1992method} &  0.7758 & 0.8129 & 0.8229 & 0.8308 & 0.8346 \\
CenterPoint~\cite{yin2021center} & 0.7270 & 0.7658 & 0.7800 & 0.7899 & 0.7935\\
S'More & \textbf{0.8323} & \textbf{0.9003} & \textbf{0.9224} & \textbf{0.9568} & \textbf{0.9582}\\
\bottomrule
\end{tabular}} 
\label{tab:latency}
\end{table}

\vspace{0.1cm}
\noindent \textbf{Loss Components.}
We ablate each loss component and report the results under various flow thresholds in \cref{tab_loss}. Notably, we observe that as the flow threshold decreases, there is a corresponding degradation in model performance. This trend aligns with our expectation, as lower thresholds are designed to detect subtler motions. 
Subtler motions often come with more severe swimming effects and thus lead to less accurate predictions.
We also observe that each loss component plays a critical role in tuning the model for these subtleties, making them essential for maintaining performance across varying motion dynamics. 

\vspace{-5pt}
\section{Conclusion}

This paper defines the problem of perceiving subtle motion for vehicles, presenting practical significance. To mitigate swimming artifacts causing ambiguity in subtle motion perception, we leverage occupancy completion as an effective strategy to facilitate motion learning. Despite the overall good performance, our method faces challenges under extremely sparse or high-occluded objects. Also, we currently only handle vehicles but not pedestrians or cyclists. We hope our work and its limitations can inspire more research into this important yet under-explored problem.

\section*{Acknowledgement}
This research project has been partially funded by research grants  to Dimitris N. Metaxas through NSF: 2310966, 2235405, 2212301, 2003874, and FA9550-23-1-0417.
{
    \small
    \bibliographystyle{ieeenat_fullname}
    \bibliography{main}
}


\end{document}